\documentclass[a4paper,fleqn]{cas-dc}

\usepackage{subcaption}

\usepackage[authoryear]{natbib}

\def\tsc#1{\csdef{#1}{\textsc{\lowercase{#1}}\xspace}}
\tsc{WGM}
\tsc{QE}

\begin{document}

\let\WriteBookmarks\relax
\def\floatpagepagefraction{1}
\def\textpagefraction{.001}

\shorttitle{Use of a low-cost forward-looking sonar for collision avoidance in small AUVs, analysis and experimental results}    

\shortauthors{C. Morency, D. J. Stilwell, S. T. Krauss,}  

\title[mode = title]{Use of a low-cost forward-looking sonar for collision avoidance in small AUVs, analysis and experimental results}

\author[1]{Christopher Morency}[type=editor, orcid=0000-0003-1362-9011]

\cormark[1]

\ead{cmorency@vt.edu}

% \credit{Conceptualization, Methodology, Software, Investigation, Formal Analysis, Validation, Writing - original draft}

\affiliation[1]{organization={Bradley Department of Electrical and Computer Engineering, Virginia Polytechnic Institute and State University},
            city={Blacksburg},
            postcode={24060}, 
            state={VA},
            country={USA}}

\author[1]{Daniel J. Stilwell}[type=editor, orcid=0000-0002-5410-2024]

% Email id of the second author
\ead{stilwell@vt.edu}

% \credit{Conceptualization, Writing – review \& editing, Supervision, Project administration, Funding Acquisition}

\author[1]{Stephen T. Krauss}[type=editor, orcid=0009-0005-0284-1879]

% Email id of the second author
\ead{stkrauss@vt.edu}

% \credit{Writing – review \& editing}

\cortext[1]{Corresponding author}

\begin{abstract}
In this paper, we seek to evaluate the effectiveness of a novel forward-looking sonar system with a limited number of beams for collision avoidance for small autonomous underwater vehicles (AUVs). We present a collision avoidance strategy specifically designed for a novel forward-looking sonar system based on posterior expected loss, explicitly coupling the obstacle detection, collision avoidance, and planning. We demonstrate the strategy with field trials using the 690 AUV, built by the Center for Marine Autonomy and Robotics at Virginia Tech, and verify the forward-looking sonar system using a prototype sonar with nine beams. Post-processed simulations are performed while changing parameters in the sensitivity of the system to demonstrate the trade-off between the detection and false alarm rates. 
\end{abstract}

% \begin{highlights}
% \item Present a coupled obstacle detection, avoidance and planning approach that is well-suited to low-cost sonar systems that have only a few sonar beams
%     \item Explicitly incorporate the uncertain information obtained from the low-cost forward-looking sonar and uncertain motion of the AUV through a Bayesian analysis, specifically minimizing the posterior expected loss for the AUV's decision control law
%     \item Demonstrate that the proposed collision avoidance method is appropriate for the proposed forward-looking sonar system with a limited number of beams through field trials
% \end{highlights}

% Keywords

\maketitle

\section{Introduction}

We seek to investigate the collision avoidance problem for autonomous underwater vehicles (AUVs) using an inexpensive sonar with an array of forward-looking beams. We approach the problem through a coupled obstacle detection, avoidance, and planning approach centered on Bayesian expected loss. We derive a collision avoidance method that incorporates uncertainty in vehicle dynamics and a probabilistic map of the environment. Our collision avoidance method balances the cost of potential collisions and the goal of reaching a desired waypoint. Through field trials, we verify that our coupled approach is well suited to a forward-looking sonar system that uses a low number of wide-angle sonar beams. We demonstrate our approach to aid in the development of a novel sonar system for collision avoidance for small to medium sized AUVs.

Collision avoidance using wide-angle, forward-looking sonars has many inherent challenges, including noisy data and angular position uncertainty in the sonar beams. Forward-looking sonars may have good distance discretization; however, obstacles appearing at the \emph{same} distance are indistinguishable within a single sonar beam. Imaging sonars, such as the DIDSON \citep{Belcher_2002} with 96 beams or the Blueview P450 Series \citep{Horner_2009} with between 256 and 768 beams can overcome these challenges by incorporating hundreds of narrow beams that minimize the angular positional uncertainty. However, current imaging sonars used for collision avoidance are prohibitively high in size, cost, and power consumption for many small AUVs. For small, low-cost AUVs, a sonar that is more capable than a single-beam forward-looking sonar \citep{Hutin_2005}, yet less expensive, smaller, and requiring less power than an imaging sonar can be beneficial for operating AUVs in cluttered environments with multiple potential obstacles. We employ a middle ground solution using nine wide-angle beams.

\begin{figure}[t]
\centering 
\includegraphics[width=\linewidth]{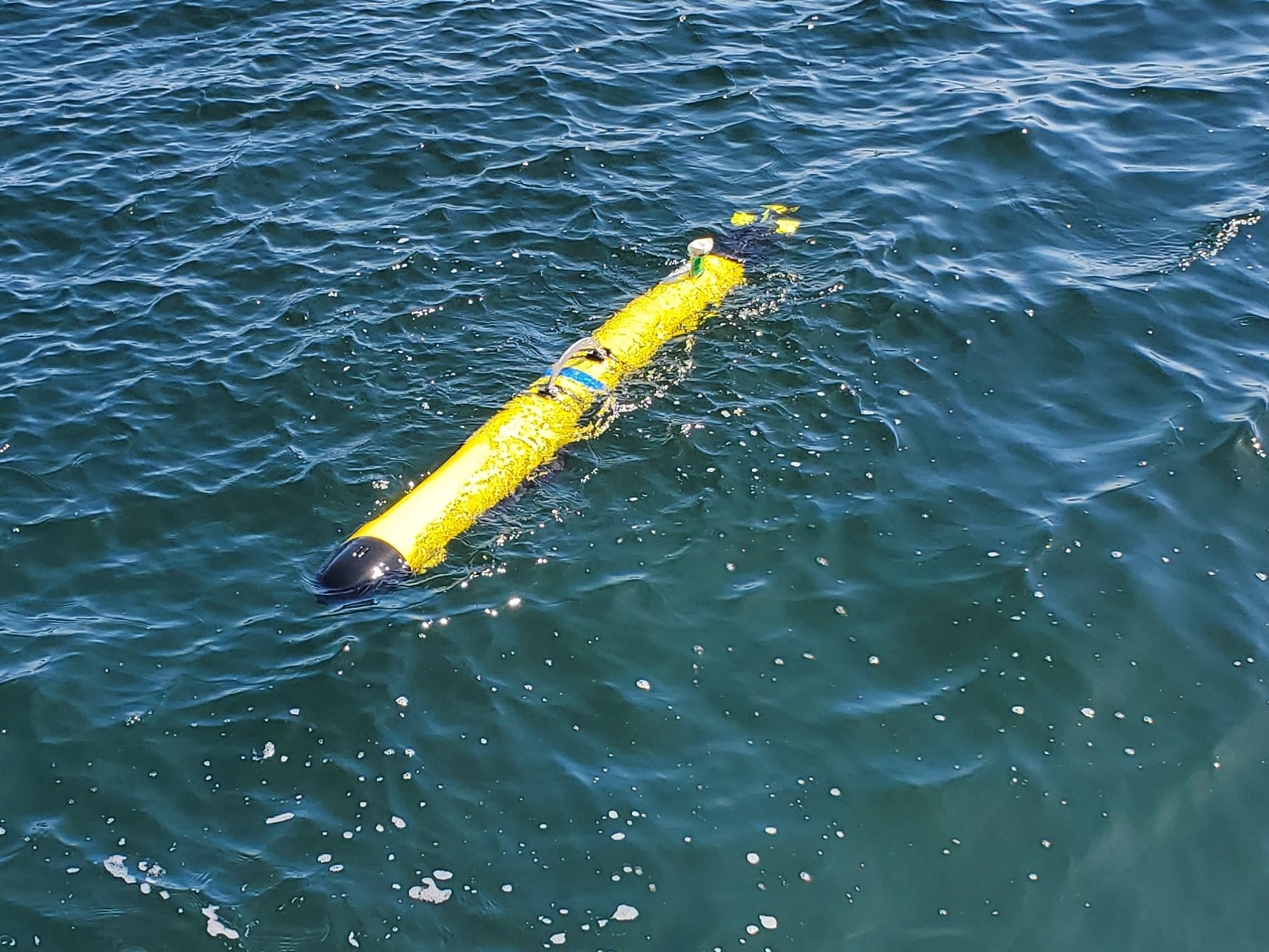}
\caption{Virginia Tech's Center for Marine Autonomy and Robotics 690 AUV}
\label{690pic}
\end{figure}

Several solutions exist for collision avoidance with single-beam sonars using an echo sounder \citep{Calado}, or using mechanically steered sonars \citep{Grefstad, Solari, Heidarsson_2011}. With a single-beam echo sounder, \citet{Calado} used concepts from histogramic in-motion mapping \citep{BorensteinHMM} to build a global obstacle map. Histogramic in-motion mapping updates a global grid map each time new sensor measurements are received by incrementing ensonified cells where an obstacle is detected and decrementing ensonified cells where an obstacle is not detected. For path planning, \citet{Calado} converted detected obstacles from the grid map into a feature based map, assuming obstacles are present at a location in the map when the value of a cell in the map is greater than a predetermined threshold. Using a mechanically scanning single-beam sonar for their Merlin ultra-compact vehicle, \citet{Grefstad} employed an occupancy grid \citep{Elfes} to map the returns from the sonar. Occupancy grids partition the environment into cells and update the cell values using a recursive Bayesian update when new measurements are received. To perform path planning, \citet{Grefstad} assumed an obstacle is present at a location in the grid if the cell value is above a predetermined threshold. In contrast, our proposed method does not use a predetermined threshold for detection or planning. We find the path with the lowest risk, incorporating the uncertainty of the sensor to find the probability of collision and deviation of the path from the waypoint.

Other methods \citep{Solari, Heidarsson_2011} using a mechanically scanning single-beam sonar have employed adaptations of artificial potential fields for obstacle avoidance. The artificial potential field method \citep{Khatib} is a common reactive approach to obstacle avoidance, which represents obstacles with repulsive forces. Researchers have proposed modifications to the potential field method \citep{Boren} such that they can react to unexpected obstacles. For vehicles with dynamic constraints such as AUVs, potential fields can result in vehicle oscillations or local minima traps. An approach to solve these issues is outlined by \citet{Boren1990} with vector field histograms; however, the method remains difficult to implement when the robot has a restrictively large turning radius \citep{Hu}. Using an adaptation of the artificial potential field with a mechanically scanning sonar, \citet{Solari} assumed obstacle detection is guaranteed, which is unrealistic given the uncertainty of the returns from a forward-looking sonar. Using the vector field histogram \citep{Borenstein1991}, the approach outlined by \citet{Heidarsson_2011} used a mechanically scanned profiling sonar on an autonomous surface vehicle and classified obstacles based on a predefined threshold. The vector field histogram employs candidate ``valleys'' to determine routes for safe travel. In contrast, our method incorporates uncertainty in obstacle detection and planning, and selects paths using a probabilistic representation of the map.

Approaches using multi-beam sonars \citep{Horner2005, Hemminger, Petillot, Teo} are abundant in the literature. Using two blazed array forward-looking sonars for vertical obstacle avoidance, \citet{Horner2005} used sonar images to build a binary map by applying a threshold to the sonar images, then searched for contours of obstacles in the binary map and sent detected obstacles to the path-planning controller. Their approach for collision avoidance used Gaussian-based additive function \citep{Hemminger} which defines the trajectory through a field of attractive and repulsive forces. The approaches outlined by \citet{Petillot} and \citet{Teo} used multi-beam forward-looking sonars with 120 and 256 beams, respectively, to detect obstacles through image processing, employing a threshold to map potential obstacles. In contrast, our method directly uses the uncertainty from the detection model to plan paths using a probabilistic representation of collision and costs for potential collisions.

Other methods using multi-beam sonars include many reinforcement learning \citep{Li, Yuan, Qiu} and genetic algorithms \citep{Alvarez, Chang, Yan} that can take advantage of the detailed data from imaging sonars. Each of the methods presented by \citet{Li, Yuan, Qiu, Alvarez, Chang, Yan} considered the planning and detection separately making these methods unsuitable when there are a low number of beams due to the high uncertainty and low resolution of wide-angle sonar beams.

There are numerous approaches to the mapping and detection problem developed for ground robots using ultrasonic sensors \citep{Hu, Boren1990, Fox}. However, underwater detection and avoidance methods have a unique set of challenges due to vehicle dynamics and limitations on sensors, making many solutions for ground robots impractical. For example, \citet{Boren1990} used a dynamic speed controller when obstacles are detected ahead of the robot, the algorithms presented by \citet{Hu} included the ability to stop or reverse, and the method outlined by \citet{Fox} considered velocities admissible only if the robot is able to stop before reaching an obstacle. Applying these approaches for ground robots to underwater robots is impractical due to the minimum forward velocity required for control and the large turning radius of most AUVs. We attempt to address many of these limitations by coupling the detection and planning to ensure our path minimizes the cost of our actions through posterior expected loss and respects the constraints on AUV motion.

We use a modified occupancy grid \citep{Elfes} for mapping to obtain a probabilistic representation of the environment in the vehicle frame. The environment is partitioned into cells, where each cell's value is modified using a recursive Bayesian update each time a new measurement is received. Variations of the standard occupancy grid have been proposed which have attempted to address some of the limitations \citep{Ganesan}, including the use of a motion model and local map. The motion model can incorporate uncertainty in the position of obstacles relative to the AUV. Using a laser rangefinder for an autonomous land vehicle, \citet{Fulgenzi} proposed a method for mapping and collision avoidance using velocity obstacles based on the Bayesian Occupancy Filter \citep{Coue}. Our approach is similar to the approach by \citet{Ganesan}; however, we specifically construct our map such that the cells match the area ensonified by the sonar, reducing the computational complexity and leading to an update only over the areas ensonified by the return. For our proposed system, an approach similar to the approach outlined by \citet{Fulgenzi} using velocity obstacles is infeasible due to the large angular positional uncertainty of objects detected by our wide-beam system.

A successful approach to the decision-making portion of the collision avoidance system should attempt to minimize false positives as well as false negatives due to noisy sensor data, which can lead to avoidable collisions if obstacles are not detected or false alarms eliminate feasible paths. There are several solutions to this issue for ground vehicles, such as Bayes' Risk, which has been proposed for collision mitigation \citep{Jansson}. Another approach outlined by \citet{Hu} minimized the Bayesian expected loss for an industrial robot obtaining the optimal decision rule by selecting the action with the minimum cost. Ground vehicles, unlike AUVs, can stop without losing control authority. To address these limitations caused by the dynamic constraints of an AUV, our method uses Bayesian expected loss, incorporating uncertainty in vehicle dynamics, and computes the probability of collision given a probabilistic trajectory. Our loss function used for the computation of the Bayesian expected loss balances the potential cost of a collision with the cost of an unnecessary maneuver or false alarm.

In our previous work \citep{Morency2}, we rigorously evaluated the benefits of multiple forward-looking sonar beams for collision avoidance for a small AUV. We investigated the benefit of adding each additional beam to a collision avoidance system to determine the trade-off between collision avoidance capabilities and the complexity of the sonar system. The sonars were evaluated through Monte Carlo simulations using a high-fidelity environmental model and simulation environment \citep{Morency}. In this work, we build upon the collision avoidance algorithms presented \citep{Morency2} by extending the algorithms from $\mathbb R^2$ to $\mathbb R^3$, coupling collision avoidance and planning, and incorporating practical elements to our algorithms for real world challenges. Additions include a path planning method using posterior expected loss and a method to weight cells closer to the AUV higher to avoid imminent collision. We verify our collision avoidance approach with field trials specifically designed to evaluate the collision avoidance algorithms. The tests use a prototype of our sonar system with five horizontal and five vertical forward-looking beams, where the center beam is common for the vertical and horizontal planes. 

Our contributions in this paper:
\begin{itemize}
    \item Present a coupled obstacle detection, avoidance and planning approach that is well-suited to low-cost sonar systems that have only a few sonar beams
    \item Explicitly incorporate the uncertain information obtained from the low-cost forward-looking sonar and uncertain motion of the AUV through a Bayesian analysis, specifically minimizing the posterior expected loss for the AUV's decision control law
    \item Demonstrate that the proposed collision avoidance method is appropriate for the proposed forward-looking sonar system with a limited number of beams through field trials
\end{itemize}
To the best of our knowledge, our coupled obstacle detection and planning algorithms are the first that rigorously apply Bayesian expected loss to the case of an AUV with a forward-looking sonar that has only a few beams.

The remainder of the paper is organized as follows: Section \ref{DM} outlines the detection and mapping model, Section \ref{CA} discusses the reactive collision avoidance and planning, Section \ref{FT} presents the field trials along with some practical considerations, and Section \ref{FT2} shows the results from changing the sensitivity of the detection algorithm through post-processing of the field trials.

\section{Detection Model}\label{DM}

In this section, we present our model for detecting potential obstacles ahead of the AUV. The model takes measurements from the sonar and creates a local polar map that contains the likelihood of an obstacle being present at various discretized distances ahead of the AUV. An AUV motion model to update the polar map is presented in Section \ref{MM} where the likelihood of an obstacle in each cell is propagated between sensor measurements. The sonar model is outlined in Section \ref{SM}.

\begin{figure}[htbp]
\centering 
\includegraphics[width=\linewidth]{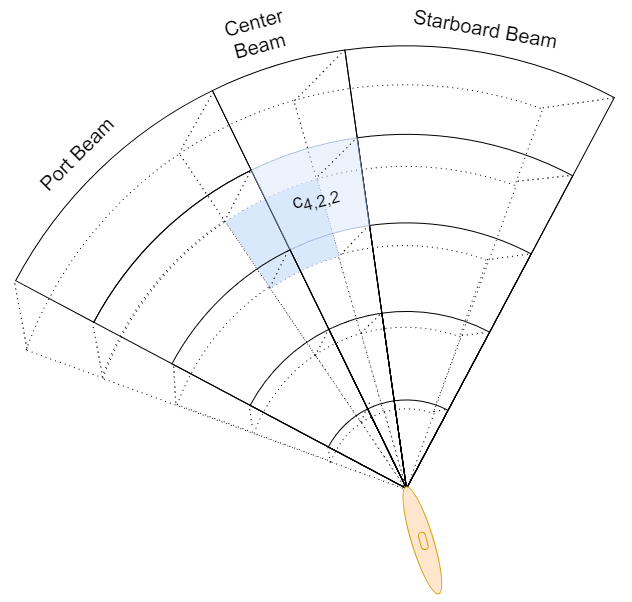}
\caption{Three-dimensional representation of horizontal polar map}
\label{polar3d}
\end{figure}

\subsection{Polar Map}\label{PM}

\begin{figure}[htbp]
\centering 
\includegraphics[width=\linewidth]{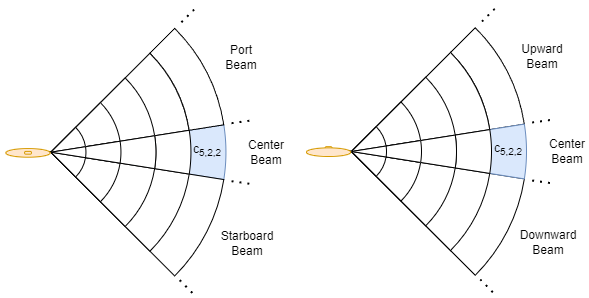}
\caption{Horizontal - top slice (left) and vertical - side slice (right) view of polar map}
\label{polar}
\end{figure}

In order to represent the the probability of an obstacle being present at a location relative to the AUV, a polar map $\mathcal M$ containing cells is constructed in an AUV-relative spherical coordinate system with the AUV at the center, as shown in Figure \ref{polar3d}. The polar map contains $m\times n \times o$ cells labeled $c_{i,j,k} \in \mathbb R^3$, where the three dimensional region defined by the cell encompasses the area of the 5dB beam pattern loss from a single return of the sonar. The cells are assigned values representing the probability of an obstacle being contained within its boundaries. Each cell $c_{i,j,k} = (r,\theta,\phi) \in \mathbb R^3$ is defined by a range of radial distances $r$, a range of azimuth angles in the horizontal axis, $\theta$, and a range of polar angles in the vertical axis, $\phi$ such that
\begin{equation*}
c_{i,j,k} = (r, \theta, \phi) :
    \begin{cases}
      r_i < r \leq r_{i+1} \\
      \theta_j < \theta \leq \theta_{j+1} \\
      \phi_k   < \phi \leq \phi_{k+1} \\
    \end{cases}
\end{equation*}
where $r_i = (i-1) l_c$, $l_c$ is the length of the cell in meters defined by the return of the sonar, and $i$ is the \textit{ith} cell ahead of the sonar. The cell $c_{i,j,k}$ is bounded between $(\theta_j, \theta_{j+1}]$ horizontal degrees and $(\phi_k, \phi_{k+1}]$ vertical degrees from the center-line of the sonar. Figure \ref{polar} shows example horizontal and vertical slices of the map, where the cell $c_{5,2,2}$ is the fifth cell in the center beam, the second cell horizontally and the second cell vertically. A 3D representation of the center layer horizontal beams is shown in Figure \ref{polar3d} where the cell $c_{4,2,2}$ is the fourth cell in the center beam and the second cell horizontally and vertically.

\subsection{Measurement Model}

When the sensor measurements are obtained, the polar map $\mathcal M$ from section \ref{PM} is updated using the measurement model. The sensor acquires a set of stochastic measurements denoted by $z^t \in \mathcal{Z}$ at time $t$. By utilizing our measurement model \citep{Morency}, we can calculate the probability of a measurement given the presence of an obstacle within a specific cell at time $t$, denoted by $c_{i,j,k}^t$, which is represented as $p(z^t|c_{i,j,k}^t = 1)$.

To incorporate newly acquired sets of measurements $z^t$ from the sonar, a recursive Bayesian update is computed for each cell $c_{i,j,k} \in \mathcal{M}$. The posterior probability that $c_{i,j,k}^{t} = 1$ given the new measurements is calculated using the following equation for the Bayesian update

\begin{equation*}
p\left(c_{i,j,k}^{t} = 1| z^t\right) = \frac{p(z^t| c_{i,j,k}^{t} = 1)p(c_{i,j,k}^{t} = 1)}{\sum_{k = 0}^1 p(z^t|c_{i,j,k}^{t} = k)p(c_{i,j,k}^{t} = k)}
\end{equation*}
Here, $p(z^t|c_{i,j,k}^{t} = 1)$ is derived from the sensor model, and $p(c_{i,j,k}^{t} = 1)$ is the prior obtained from the propagated probabilities of the velocity distributions, described in Section \ref{MM}. The probability that there is no obstacle in a given cell is denoted by $p(c_{i,j,k}^{t} = 0)$, which is equivalent to $1 - p(c_{i,j,k}^{t} = 1)$.

\subsection{Motion Model}\label{MM}

The motion model propagates the polar map between sensor measurements by leveraging the relative rotational and translational velocity of the AUV in relation to the environment. To improve real-time performance, the rotational and translational motion are separated. We update the cells based on a distribution of possible velocities relative to the AUV, where each cells $c_{i, j, k}$ utilizes the same distribution for the AUV's motion. 

To update the polar map, the probability of an obstacle in each cell is computed, taking into account the elapsed time $\tau$ between measurements and the relative translational and rotational velocities $\tilde v$ between sensor measurements. Specifically, the probability of an obstacle in cell $c_{i,j,k}$ at time $t+\tau$ is given by

\begin{equation}\label{relative velocity eq}
\begin{split}
            p(c_{i,j,k}^{t+\tau} = 1) = & \bigcup_{m = 1}^{M}\bigcup_{n = 1}^{N} \bigcup_{o=1}^O \bigg[p(c_{m,n,o}^{t} = 1) \\ &\int_{\tilde{v}\in \mathcal V} \frac{V(c_{i, j, k},c_{m, n, o}|\tilde{v}', \tau)}{V(c_{i,j,k}) } p(\tilde{v})d\tilde{v}'\bigg] 
\end{split}
\end{equation}
Here, $V\left(c_{i, j, k},c_{m, n, o}| \tilde v, \tau\right)$ represents the overlapping volume after a displacement over time $\tau$ of cells $c_{i,j,k}$ and $c_{m,n,o}$ given a relative velocity $\tilde v$. $V(c_{m,n,o})$ denotes the total volume of cell $c_{m, n, o}$, and $\mathcal V$ is the distribution for velocity.

A constant velocity distribution is assumed and each cell is independent, which allows the unions in equation \eqref{relative velocity eq} to be expressed as sums, resulting in

\begin{equation}\label{rel vel 2}
\begin{split}
        p(c_{i,j,k}^{t+\tau} = 1) = &\sum_{m = 1}^{M}\sum_{n = 1}^{N}\sum_{o=1}^O \bigg[p(c_{m,n,o}^{t} = 1) \\ &\int_{\tilde{v}\in \mathcal V} \frac{V(c_{i, j, k},c_{m, n, o}|\tilde{v}', \tau)}{V(c_{i,j,k}) } p(\tilde{v})d\tilde{v}'\bigg]
\end{split}
\end{equation}

\begin{figure*}[htbp]
\centering 
\includegraphics[width=0.8\linewidth]{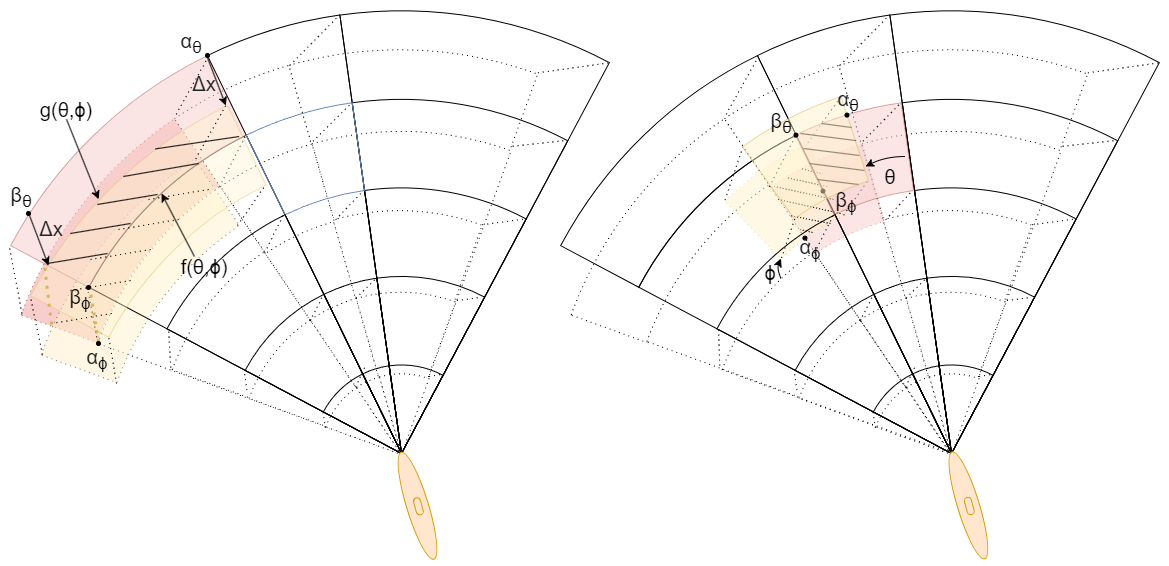}
\caption{Translational propagation (left) and rotational propagation (right)}
\label{prop3d}
\end{figure*}

To enable real-time computation, the translational and rotational motion are decoupled, and equation \eqref{rel vel 2} is first calculated with $\mathcal V$ representing the distribution for relative translational motion. For a given translational velocity $v_x$, the overlapping volume of two cells is determined using

\begin{equation*}
\begin{split}
    V&\left(c_{i, j, k}, c_{m, n, o} | v_x, \tau \right) \\ =& \int_{\alpha_\phi}^{\beta_\phi} \int_{\alpha_\theta}^{\beta_\theta} \int_{f(\theta, \phi)}^{g(\theta, \phi)} r^2 \sin(\phi)  dr   d\theta d\phi
\end{split}
\end{equation*}
where $f(\theta, \phi)$ and $g(\theta, \phi)$ represent the horizontal bounds of the overlapping cell. $\alpha_\theta$ and $\beta_\theta$ define the minimum and maximum horizontal angles, respectively, and $\alpha_\phi$ and $\beta_\phi$ are the minimum and maximum vertical angles, respectively, for which a point in cell $c_{m,n,o}$ overlaps with cell $c_{i,j, k}$. An illustration of the translational propagation is shown in Figure \ref{prop3d} on the left, where the red cell is shifted by a displacement of $\Delta x = v_x\tau$, the yellow is the cell after the displacement, and the overlappLaing area of the cell onto itself is represented by the shaded area with horizontal lines.

The relative rotational motion, $v_\omega$, is further split into $v_\theta$ and $v_\phi$ which are the horizontal and vertical components of the relative rotational motion, respectively. Equation \eqref{rel vel 2} is computed for the relative rotational motion, where $\mathcal V$ now represents the distribution of possible rotational velocities relative to the AUV. The overlapping volume, $V(c_{i, j, k},c_{m, n, o}|v_\omega, \tau)$, is computed for the rotational velocities $v_\theta$ and $v_\phi$, over the time $\tau$. For a given $v_\theta$ and $v_\phi$, the overlapping volume depends on whether the cells are the same distance from the sonar

\begin{equation*}
    V\left(c_{i, j, k},c_{m, n, o}|\omega_\theta, \omega_\phi, \tau\right) = \begin{cases} \frac{\beta_\theta - \alpha_\theta}{|\theta_{n+1} -\theta_n|} \frac{\beta_\phi-\alpha_\phi}{|\phi_{o+1} - \phi_o|}  & i = m \\ 0 & i \neq m\end{cases}
\end{equation*}
Here, $\alpha_\theta$ and $\beta_\theta$ represent the minimum and maximum horizontal overlap angles, $\alpha_\phi$ and $\beta_\phi$ represent the minimum and maximum vertical overlap angles. $\theta_{n+1}$ and $\theta_n$ are the horizontal bounds of cell $c_{m,n,o}$ while $\phi_{o+1}$ and $\phi_o$ are the cell's vertical bounds. The rotational propagation is illustrated in Figure \ref{prop3d} on the right, where the cell is shifted by a horizontal rotation of $\theta$ and a vertical rotation of $\phi$ and the shaded area with horizontal lines represents the overlapping area. 

If navigation data is available, it can be used to choose the translational and rotational velocity distributions. Otherwise, a uniform distribution of possible velocities or a normal distribution with a large enough variance to capture the potential velocities should be selected.

\subsection{Sonar Model}\label{SM}

To simulate the acoustic propagation of the sonar, we use our high-fidelity sonar model \citep{Morency}. The model involves a set of equations for various parameters such as sound velocity \citep{Medwin_1975}, attenuation \citep{z1}, \citep{z2}, spread loss \citep{Jenson_2011}, beam pattern loss through the single-point-source method \citep{Marage_2010}, backscatter \citep{Urick_1983}, sonar resolution \citep{Medwin_1998}, and noise \citep{Coates_1990}. The backscatter computation is performed for the seafloor, sea surface, and volume. In addition, the noise sources considered include turbulence, shipping traffic, sea state, and thermal noise.

\section{Collision Avoidance}\label{CA}

This section introduces a reactive collision avoidance approach, which is divided into three parts. In Section \ref{CA1}, a loss function and decision rule are presented for determining when the AUV should react to an obstacle. Section \ref{CA2} extends this standard approach to incorporate a probabilistic trajectory. In Section \ref{CA3}, we introduce a path planning approach that is coupled to the framework described in Section \ref{CA1}.

Our method employs the posterior expected loss, which uses the polar map we constructed and employs a loss function defined for each available maneuver to compute the optimal decision function. This framework is based on the standard decision-theory framework proposed by \citet{Hu} and \citet{Jansson}.

\subsection{Decision Rule}\label{CA1}

Let $\mathcal{A}$ denote the set of actions available to the AUV, where each action is represented by $a^k \in \mathcal{A}$. The action space can be as simple as a decision on whether to intervene, as in the case of the pencil beam sonar, or it can consist of multiple possible actions, such as the five actions represented by $\mathcal{A}$

\begin{equation*}
\mathcal{A} = \begin{cases} a^0 & \text{go straight} \\ a^1 & \text{turn left} \\ a^2 & \text{turn right} \\ a^3 & \text{go up} \\ a^4 & \text{go down} \end{cases}
\end{equation*}
The posterior expected loss associated with taking action $a^k$ is given by

\begin{equation*}
R(a^k| z_{0:t}) = \sum_{j = 0}^1 L(a^k, H_j)P(H_j| z_{0:t})
\end{equation*}
where $L(a^k, H_j)$ is the loss incurred by taking action $a^k$ in the event that hypothesis $H_j$ is true, $j=1$, or false $j=0$, and $P(H_j| z_{0:t})$ is the posterior probability of hypothesis $H_j$ given the measurements $z_{0:t}$. For our application, $H_1$ indicates a collision and $H_0$ indicates no collision. The decision function is modified to select the action $a^*$ that minimizes the overall posterior expected loss

\begin{equation}\label{PEL}
a^* = \underset{a^i}{\text{argmin }} R(a^i| z_{0:t}) = \underset{a^k}{\text{argmin}} \sum_{j = 0}^1 L(a^k, H_j)P(H_j| z_{0:t})
\end{equation}

The loss function is responsible for balancing the potential cost of a collision with the cost of an unnecessary maneuver or false alarm. For a given action $a^k$, the loss when $H_0$ occurs is represented by $C_0^k$, which is defined as the cost of taking the trajectory associated with action $k$. The loss when $H_1$ occurs is $C_0^k + C_1^k$, where $C_1^k$ represents the cost of collision associated with action $k$. It is important to note that $C^k_{1}$ is equal to $C^j_{1}$ for all $j$ and $k$, since the cost of a collision is the same for each action.

To account for the cost of false alarms, the cost is modified to be $C_{0}^k - C_0^0$ for all $k \neq 0$, where $C_0^0$ is the cost of not taking any avoidance maneuver and continuing on the path from the path planner. The cost of a missed detection, which would result in a collision, is $C_{1}^0 + C_0^0$. By modifying the loss function in this way, the decision-making process can be tuned to better balance the cost of a collision with the cost of a false alarm.

\subsection{Probabilistic Trajectory}\label{CA2}

We adopt a stochastic approach to AUV trajectory planning. This approach considers the probability of collision by utilizing a decision function that estimates the likelihood of a collision based on the set of possible commands and each possible trajectory. To model the AUV dynamics stochastically, we consider a finite set of candidate trajectories $X_{t:t+T} \subset \mathcal X$, where $\mathcal X$ is the set of all possible trajectories, and the probability of each trajectory is determined for every possible command. By computing the probability of possible trajectories given a maneuvering command and accounting for the probability of obstacles occupying each cell, we arrive at the likelihood of collision for a given command. An indicator function $I(X^k_{t:t+T} \in c_{i, j,k})$ is used to determine whether a given trajectory $X^k_{t:t+T} \in X_{t:t+T}$ passes through the cell $c_{i, j, k}$, where $T$ is the length of the trajectory in seconds, and $I(X^k_{t:t+T} \in c_{i, j,k}) = 1$ if the trajectory passes through the cell and $I(X^k_{t:t+T} \in c_{i, j,k}) = 0$ if the trajectory does not pass through the cell. Over a deterministic trajectory, the probability of collision with a cell $c_{i,j,k}$ is

\begin{equation}\label{cellcol}
    p(C(c_{i,j,k}) = 1|X^k_{t:t+T}) = I(X^k_{t:t+T} \in c_{i, j, k})p(c_{i, j, k} = 1)
\end{equation}
The probability of no collision with a cell $c_{i,j,k}$ given a trajectory $X^k_{t:t+T}$ is denoted by $p(C(c_{i,j,k}) = 0|X^k_{t:t+T})$, which is equivalent to $1 - p(C(c_{i,j,k}) = 1|X^k_{t:t+T})$. Since the cells $c_{i, j, k}$ are independent, the probability of collision over a trajectory is  

\begin{equation}\label{probcolovertraj}
\begin{split}
    p(C(X^k_{t:t+T}) = 1) = 1 - \prod_{c_{i,j,k}\in \mathcal M} p(C(c_{i,j,k}) = 0|X^k_{t:t+T})
\end{split}
\end{equation}
To incorporate probabilistic trajectories, the probability that an action $a^k$ results in collision is

\begin{equation}\label{pcol traj}
    p(C(X^k_{t:t+T}) = 1| a^k)   = \int_{X^k_{t:t+T} \in \mathcal{X}}p(x|a^k)p(C(x) = 1)dx
\end{equation}
where $p(x|a^k)$ is the probability of trajectory $x = X^k_{t:t+T} \in \mathcal X$ given an action $a^k$ and $p(C(x) = 1)$ is the probability of a collision resulting from a trajectory $x$ from equation \eqref{probcolovertraj}.

We employ a receding horizon \citep{Biggs} approach for planning. Receding horizon planning computes a path over a finite planning horizon and executes a portion of that path before re-planning. For collision avoidance, the planning horizon is much longer than the time before the algorithm re-plans since a new path is computed after every new set of measurements from the sonar $z^t$. Therefore, when obstacles are far from the sonar, the algorithm will have many opportunities to recompute the path before reaching the obstacle whereas when obstacles are close to the sonar, the system will have fewer opportunities to compute new paths. To account for the many re-computations due to the nature of receding horizon planning, we consider imminent collisions with a greater weight than further collisions along the trajectory. We modify equation \eqref{cellcol} to be the cost of a trajectory given an action $a^k$ such that 

\begin{equation}\label{weightedcost}
\begin{split}
    C(C(&c_{i,j,k}) = 1|X^k_{t:t+T}) = \\ &I(X^k_{t:t+T} \in c_{i, j, k})p(c_{i, j, k} = 1)\frac{I_x-(i-1)c_d}{I_x} 
\end{split}
\end{equation}
where $I_x$ is the furthest cell index away from the sonar and $c_d\in [0, 1]$ is strategically chosen to balance the cost of avoiding imminent collisions and the cost of planning for future potential collisions. When $c_d = 0$, every cell is weighted equally. If $0 < c_d \leq 1$, the closest cells to the sonar are weighted the highest and each subsequent cell is weighted lower as the distance from the sonar increases. Employing this updated cost function, and using $C(C(c_{i,j,k}) = 1|X^k_{t:t+T})$ in equation \eqref{probcolovertraj}, our modified decision function becomes

\begin{equation*}
    a^* = \underset{a^k}{\text{argmin}} \sum_{j = 0}^1 L(a^k, H_j) C(C(X_{t:t+T}) = 1| a^k)
\end{equation*}

\begin{figure*}[t]
\centering
\begin{subfigure}{.5\textwidth}
    \centering
    \includegraphics[width=\linewidth]{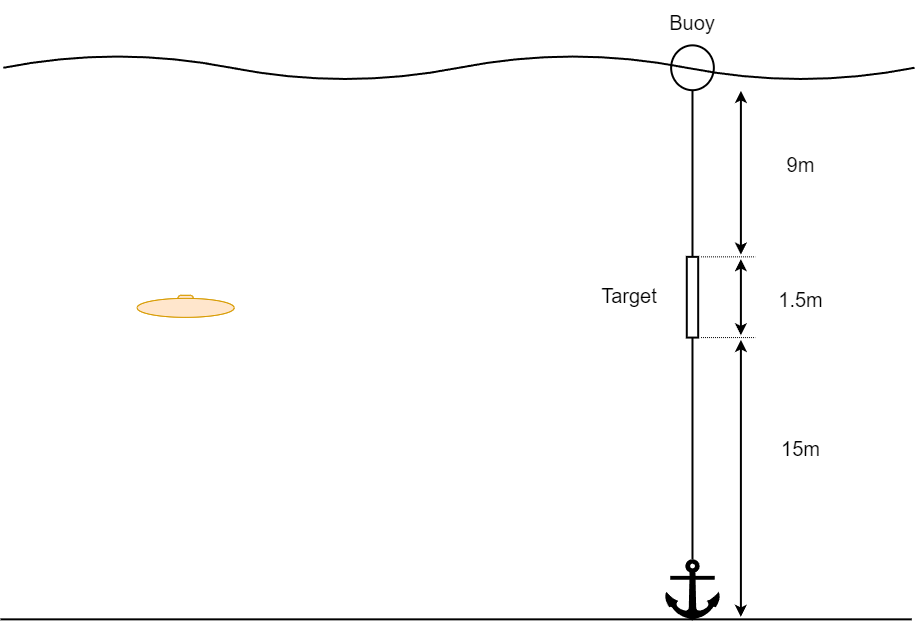}
    \caption{Target setup}
    \label{experiment}
\end{subfigure}%
\begin{subfigure}{.5\textwidth}
    \centering
    \includegraphics[width=\linewidth]{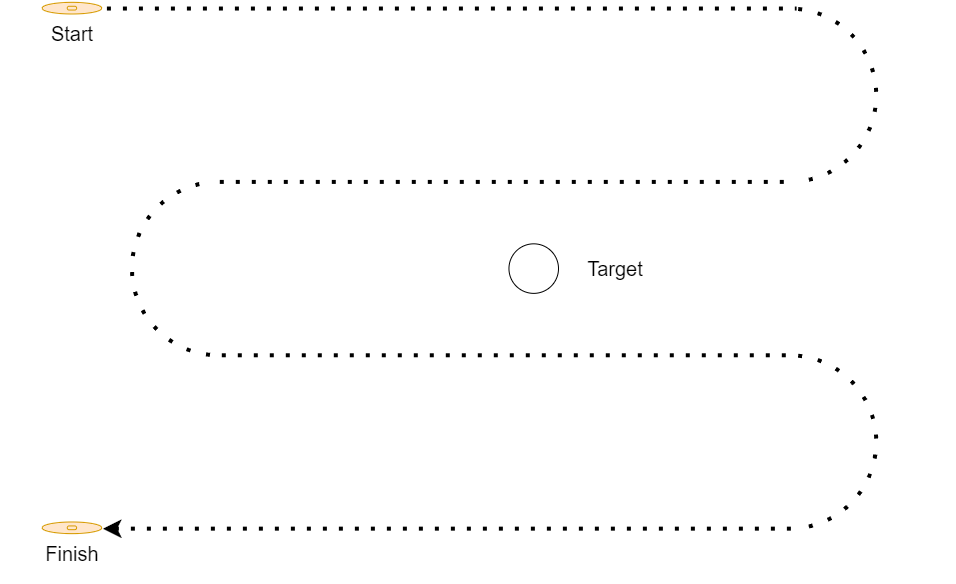}
    \caption{Lawnmower path}
    \label{lawnmower2}
\end{subfigure}
\caption{Experimental setup for collision avoidance field trials}
\label{Experimental}
\end{figure*}

\subsection{Path Planning}\label{CA3}

An integrated path planner is needed to incorporate the collision avoidance algorithm with the high-level navigation and goals of the AUV. To couple the AUV's path planner with our collision avoidance decision rule, we extend the loss function in section \ref{CA1} by including costs for deviation from the AUV's path. To achieve this, we consider a simple path planning algorithm using waypoints and the posterior expected loss \eqref{PEL} discussed in Section \ref{CA1}. The waypoints are generated by the 690 AUV's path planner and the collision avoidance system takes control when our system detects that the probability of a collision along the current trajectory is greater than our predetermined risk tolerance. Our method computes the current yaw and pitch angles, and measures the difference between the angle the AUV is facing and the desired angle required to reach the waypoint for each possible trajectory we evaluate. This is achieved by constructing the cost of each $a^k$ in \eqref{PEL} for each of the possible trajectories such that 

\begin{equation*}
\begin{split}
     &C_0^k = \cos^{-1}\bigg[\cos(\phi_w)\cos(\phi_k)\\ &\big(\cos(\theta_w)\cos(\theta_k) + \sin(\theta_w)\sin(\theta_k)\big) + \sin(\phi_w)\sin(\phi_k)\bigg] c
\end{split}
\end{equation*}
where $\theta_k$ and $\phi_k$ are the final heading and pitch angles as a result of taking trajectory $k$; $\theta_w$ and $\phi_w$ are the heading and pitch angles to the waypoint. This results in the angle between the forward vector of the AUV and the vector to the waypoint multiplied by a user selected cost labeled by a constant $c$. The constant $c$ is strategically chosen to balance the cost of maintaining the desired trajectory in the AUV mission and the cost of avoiding obstacles. A lower value of $c$ will result in maintaining the vehicle path, while a high value of $c$ will minimize the probability of collision. Incorporating the weighted costs for cells in Section \ref{CA2} and the updated costs for path planning, the resulting action $a^*$ from \eqref{PEL} becomes 

\begin{equation*}
\begin{split}
    a^* = \underset{a^k}{\text{argmin}} \bigg[ &C_0^kC(C(X_{t:t+T}) = 0| a^k) + \\ &(C_0^k + C_1^k)C(C(X_{t:t+T}) = 1| a^k)\bigg]
\end{split}
\end{equation*}

\section{Field Trials}\label{FT}

We conduct experiments to validate the collision avoidance algorithms and forward-looking sonar for an autonomous underwater vehicle. We use a prototype sonar consisting of five horizontal beams and five vertical beams with a common center beam. There are 219 distance bins for each beam, reaching 50 meters ahead of the sonar. The center beam has a 5 dB beam width of 10 degrees and the outer beams have beam widths of 20 degrees, centered at $\pm 15$ and $\pm 35$ degrees, on both the horizontal and vertical axis. The experiments are conducted using the Virginia Tech Center for Marine Autonomy and Robotics 690 AUV, shown in Figure \ref{690pic}, and are performed in Claytor Lake, Virginia. The experimental setup consists of a buoyant pool noodle attached to an anchor at the lake floor and a buoy at the surface, as shown in Figure \ref{experiment}. The rope is buoyant to minimize the drift of the pool noodle, which acts as the target. The experiments are conducted at a depth of 10 meters below the surface and the target is 15 meters above the lake floor. The AUV performs a lawnmower path around the target as shown in Figure \ref{lawnmower2}. The position of the target is estimated to be in the center of the lawnmower path; however, the target can drift due to external factors such as wind or wake from small boat traffic in Claytor Lake. Multiple runs of the experiment are performed using the collision avoidance algorithms described in Section \ref{CA}. The vehicle's onboard processor is an ODROID XU4 \citep{Odroid}, and is used to construct the polar map with sonar data and compute avoidance maneuvers for collision avoidance.

\begin{figure*}[htbp]
\centering
\begin{subfigure}{.333\textwidth}
    \centering
    \includegraphics[width=\linewidth]{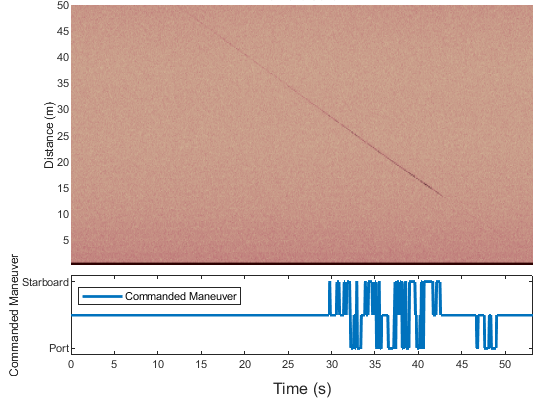}
    \caption{Forward sonar}
    \label{Forward}
\end{subfigure}%
\begin{subfigure}{.333\textwidth}
    \centering
    \includegraphics[width=\linewidth]{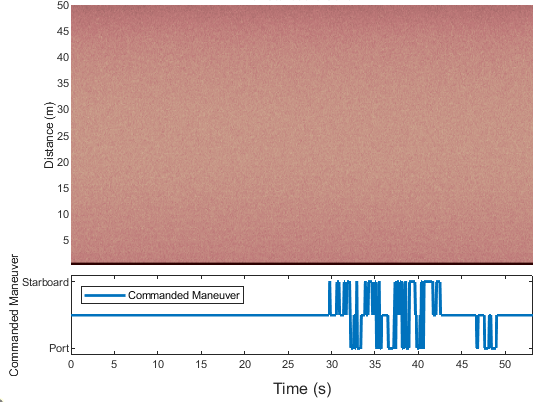}
    \caption{Far starboard sonar}
    \label{Far Starboard}
\end{subfigure}%
\begin{subfigure}{.333\textwidth}
    \centering
    \includegraphics[width=\linewidth]{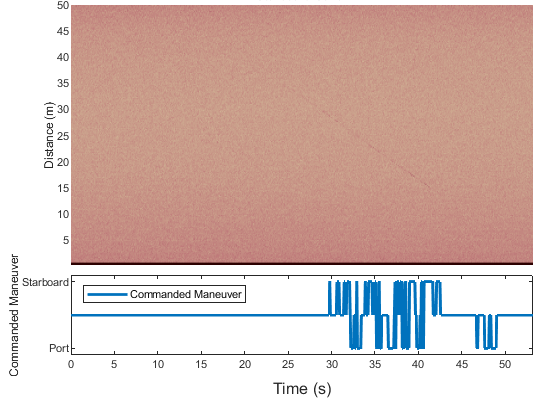}
    \caption{Starboard sonar}
    \label{Starboard}
\end{subfigure}
\begin{subfigure}{.333\textwidth}
    \centering
    \includegraphics[width=\linewidth]{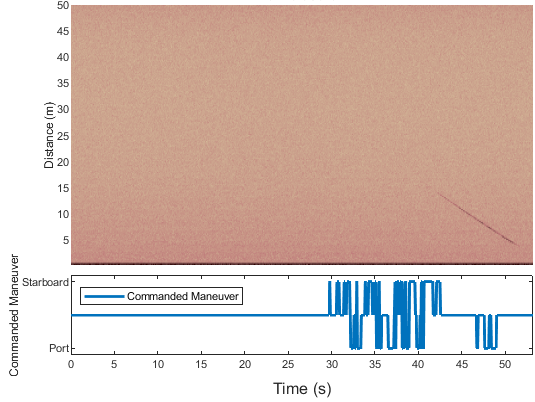}
    \caption{Far port sonar}
    \label{Far Port}
    \end{subfigure}%
    \begin{subfigure}{.333\textwidth}
    \centering
    \includegraphics[width=\linewidth]{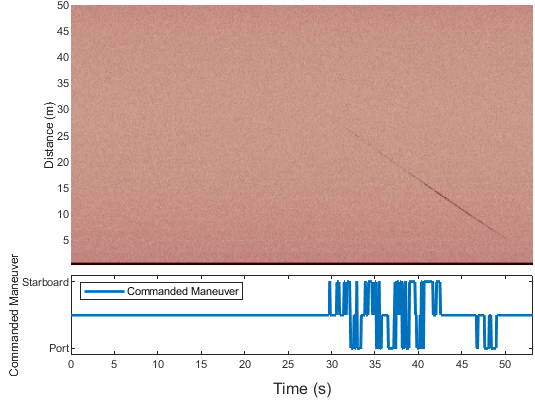}
    \caption{Port sonar}
    \label{Port}
\end{subfigure}%
\begin{subfigure}{.333\textwidth}
    \centering
    \includegraphics[width=\linewidth]{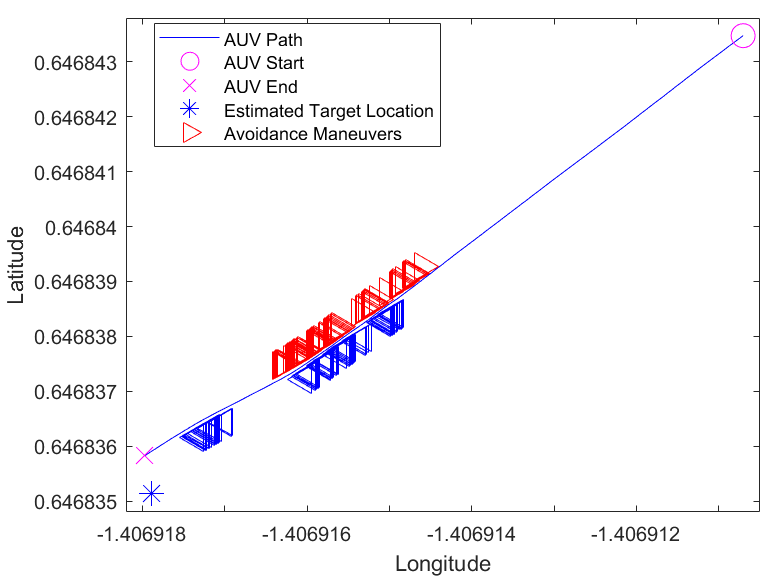}
    \caption{AUV path and commanded maneuvers}
    \label{Path}
    \end{subfigure}
\caption{Sonar returns, path and avoidance commands during field trial run in Claytor Lake}
\label{field}
\end{figure*}

\subsection{Results}\label{FT1}

The real-world experiment was performed five times, each of which successfully avoided collision using a low-cost sonar. The results from an illustrative run is presented in this section. Figure \ref{field} shows the portion of the run as the AUV approaches the target. The returns from each of the five horizontal beams are shown in Figure \ref{Forward} through Figure \ref{Port}. The vertical axis is the distance corresponding to a return, which is discretized into bins, or equivalently the discrete time of the return. The horizontal axis shows the time of the return relative to the start of the section shown, discretized into pings, where each ping is received approximately every 0.1 seconds. The intensity of the received data is normalized over the distance and represented by a color, where darker colors represent a higher intensity of energy received by the sonar. The avoidance maneuvers are shown below each plot, where each ping has a corresponding port, starboard or no avoidance command, which are labeled on the vertical axis.

The obstacle can be seen in the beam returns in Figures \ref{Forward}, \ref{Starboard}, \ref{Far Port}, and \ref{Port} as the dark diagonal line starting at around 13 seconds, which is in every horizontal beam other than the far starboard beam. The intensity of the return is initially highest in the forward and port sonar beams.

Figure \ref{Path} shows the vehicle path over the section of the mission along with the commanded maneuvers. The start of the mission is denoted by a purple circle, the end of the mission by a purple 'X', the vehicle path is the blue line and the commanded maneuvers are denoted by a triangles pointed in the direction of the commanded maneuvers, above and below the path. The starboard maneuvers are in red and the port maneuvers are blue. The estimated position of the target is represented by a blue asterisk.

As the vehicle approaches the obstacle, the vehicle begins with a starboard maneuver; however, the sonar returns reflected from the obstacle are not high enough to be registered as an obstacle in the port beams, especially as the vehicle turns away from the obstacle. This leads to the vehicle switching between starboard and port maneuvers until it reaches approximately 20 meters from the obstacle, where the sound intensity reflected back towards the sonar increases and the decision rule correctly maneuvers the AUV around the obstacle. It is important to consider that the obstacle was a pool noodle with a low target intensity, which the AUV was able to successfully avoid. If the target intensity were greater, or the sensitivity of the algorithm is increased, as shown in Section \ref{FT2}, the avoidance algorithms would command the AUV to maneuver sooner and further from the obstacle before attempting to turn back. The port maneuvers near the end of the mission show the effects of the path planner presented in section \ref{CA3} returning the AUV towards the waypoint. 

The results from the field trials show the successful operation of the coupled detection, avoidance and planning algorithms with a small, low target intensity obstacle. Depending on the application, changing the costs associated with reaching the waypoint and adjusting the sensitivity of the detection algorithm can allow an AUV operator to determine an appropriate risk tolerance for the mission, which can change the proximity of the AUV's trajectory to potential obstacles.

\begin{figure*}[h]
\centering
\begin{subfigure}{.5\textwidth}
    \centering
    \includegraphics[width=0.95\linewidth]{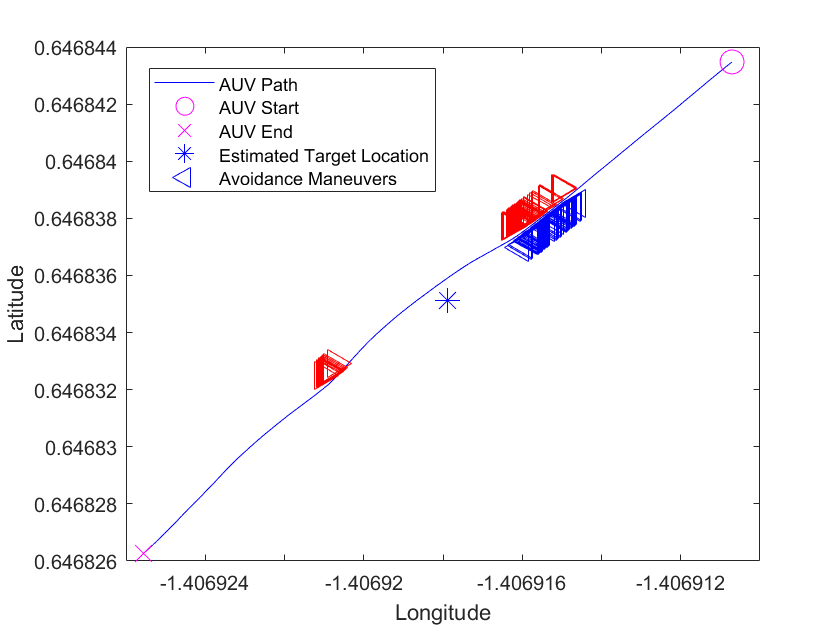}
    \caption{Low sensitivity}
    \label{PPPA}
\end{subfigure}%
\begin{subfigure}{.5\textwidth}
    \centering
    \includegraphics[width=0.95\linewidth]{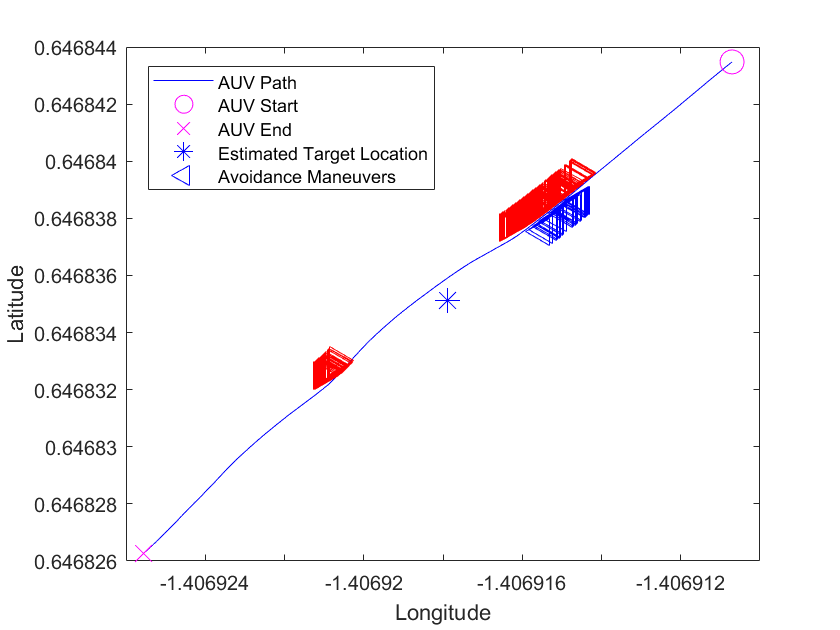}
    \caption{Medium sensitivity}
    \label{PPPB}
\end{subfigure}
\begin{subfigure}{.5\textwidth}
    \centering
    \includegraphics[width=0.95\linewidth]{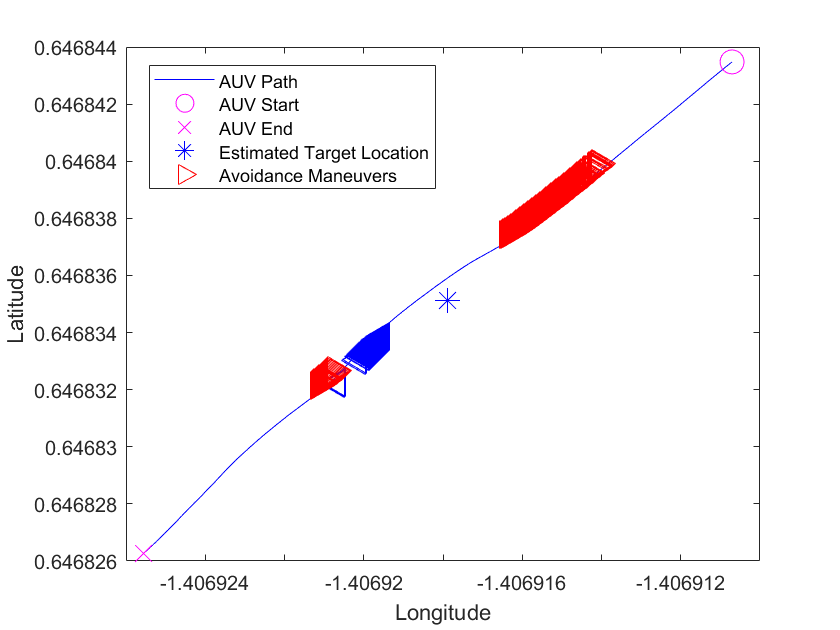}
    \caption{High sensitivity}
    \label{PPPC}
\end{subfigure}%
\begin{subfigure}{.5\textwidth}
    \centering
    \includegraphics[width=0.95\linewidth]{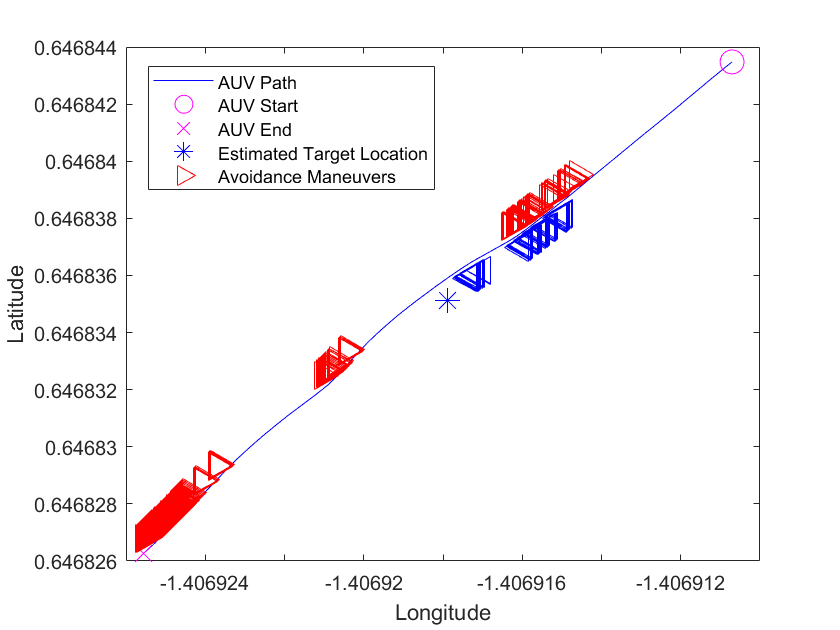}
    \caption{Real trial}
    \label{PPPD}
    \end{subfigure}
\caption{AUV path and avoidance commands at varying sensitivity settings}
\label{Post Processed Path}
\end{figure*}

\subsection{Discussion on Computation and Practical Considerations}

The computational complexity of the algorithms is a concern, especially when computing the polar map in 3D and propagating the obstacles. A practical solution to the 3D problem is to approximate the 3D polar map by computing two 2D polar maps. This can be achieved by having a single horizontal and single vertical polar map. The translational and rotational propagation of the polar map can be computationally expensive. The solution we found most practical was to limit the number of steps for the distributions of velocities and pre-computing any intensive calculations, such as the size of the bins and the overlapping areas of cells for given velocities.

One of the issues we encountered was reacting to obstacles with high target strength which were further away than an obstacle with low target strength immediately ahead, leading to maneuvers that would result in collision with a closer target. Although the algorithm is working as intended to take the path with the lowest probability of collision, this does not consider the possibility of future maneuvers that may be able to avoid the target. To resolve this potential issue, choosing a sufficiently high value for $c_d$ in equation \eqref{weightedcost} is important. 

Boat noise was a major concern for the sonar, as the false alarm rate was considerably high. Since our algorithm incorporates the path planner from Section \ref{CA3}, boat noise would result in a false alarm and minor deviation from the path, rather than a failed mission. Future iterations of the algorithm could investigate methods to identify and remove boat noise in the sonar data.

\begin{figure*}[h]
\centering
\begin{subfigure}{\textwidth}
    \centering
    \includegraphics[width=.8\linewidth]{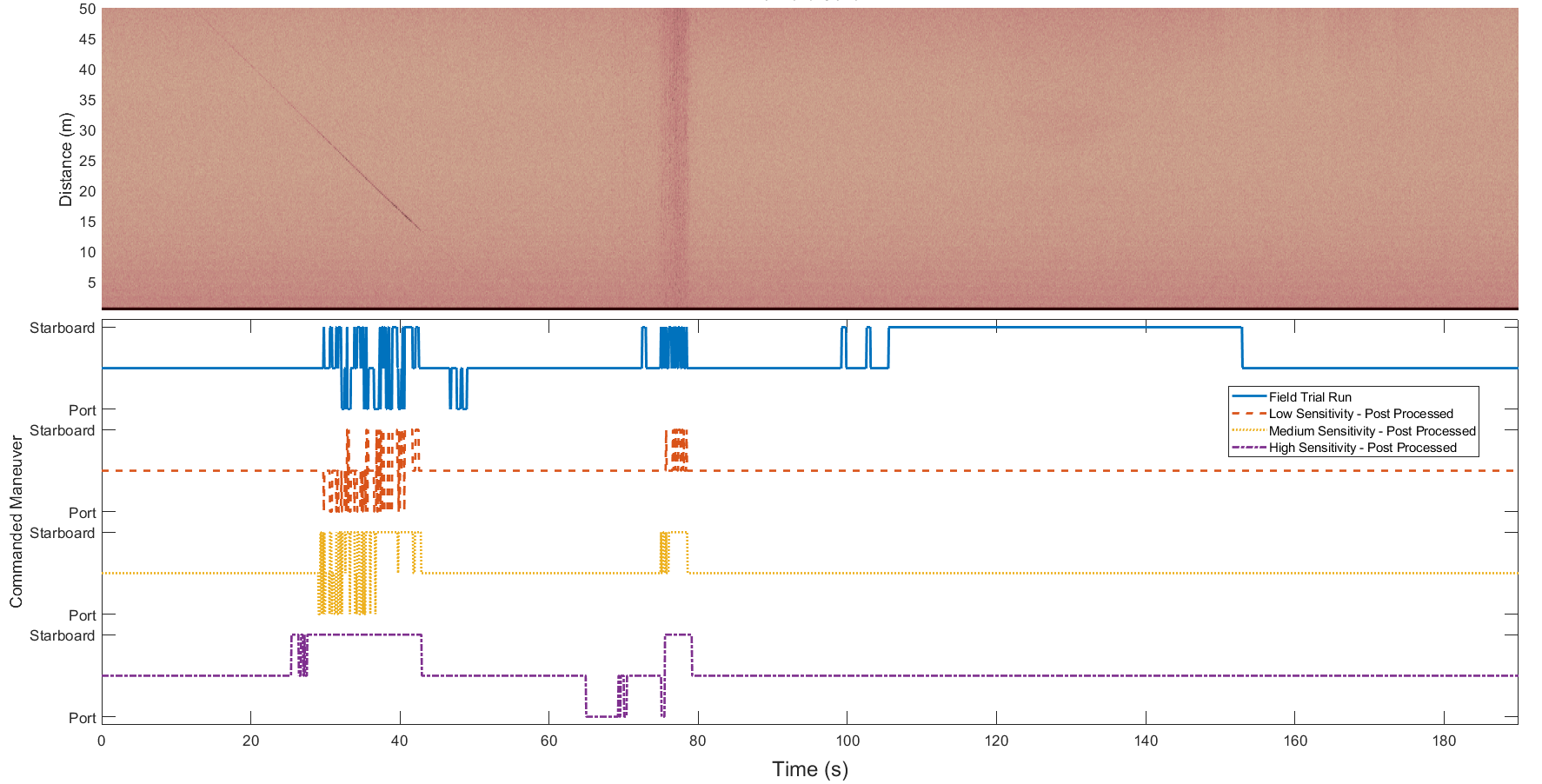}
    \caption{Returns from forward beam and commanded maneuvers from entire leg of lawnmower path}
    \label{}
\label{ReturnsA}
\end{subfigure}
\begin{subfigure}{\textwidth}
    \centering
    \includegraphics[width=.8\linewidth]{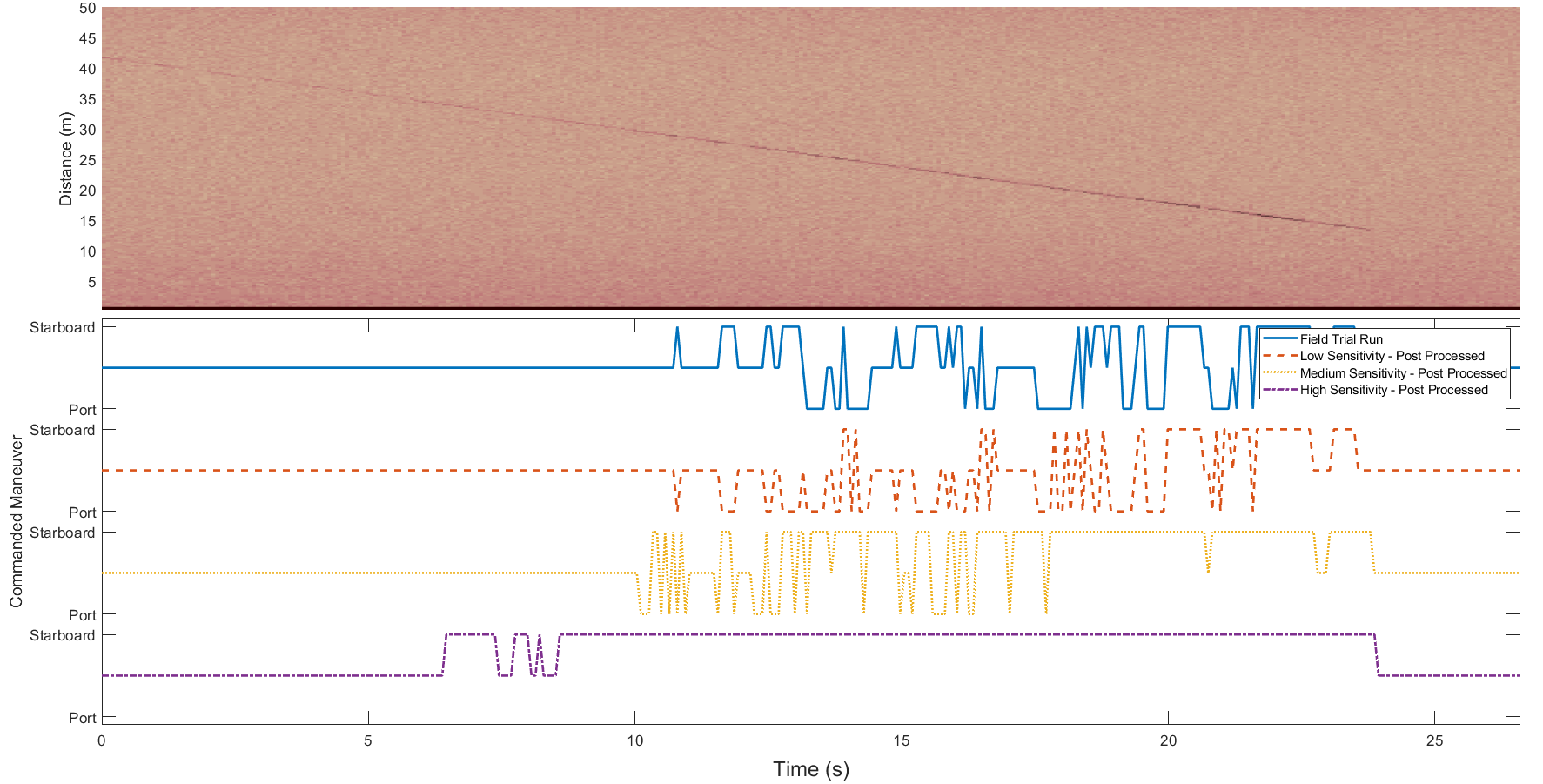}
    \caption{Close-up of returns from forward beam and commanded maneuvers at the target}
\label{ReturnsB}
\end{subfigure}
\caption{Forward beam returns as AUV approaches the obstacle with maneuvering commands shown}
\label{Post Processed Returns}
\end{figure*}

\begin{figure*}[htbp]
\centering
\begin{subfigure}{.333\textwidth}
    \centering
    \includegraphics[width=\linewidth]{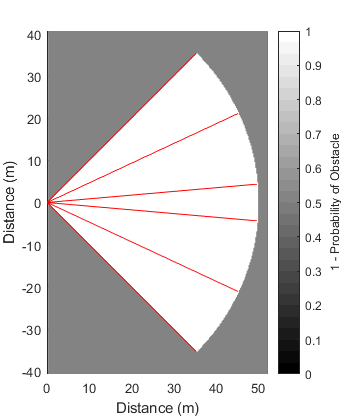}
    \caption{Polar map with low sensitivity}
    \label{local10}
\end{subfigure}%
\begin{subfigure}{.333\textwidth}
    \centering
    \includegraphics[width=\linewidth]{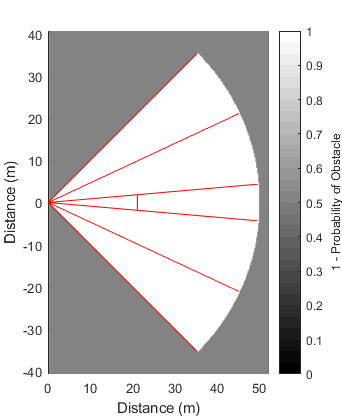}
    \caption{Polar map with medium sensitivity}
    \label{local7}
\end{subfigure}%
\begin{subfigure}{.333\textwidth}
    \centering
    \includegraphics[width=\linewidth]{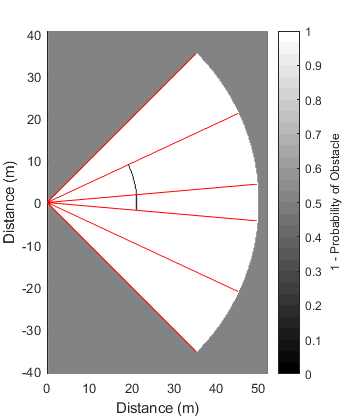}
    \caption{Polar map with high sensitivity}
    \label{local4}
\end{subfigure}
\caption{Post-processed local polar maps of sonar returns at 36 seconds for varying sonar sensitivity levels}
\label{localpost}
\end{figure*}

\section{Post Processed Field Trials}\label{FT2}
The returns from the sonar in the scenario presented in Section \ref{FT} are post-processed to evaluate the algorithm's performance. The sensitivity of the detection algorithm was systematically varied to investigate its impact on the detection performance.  The simulations are performed for three sensitivity levels by changing the expected returns of the obstacle. The three levels of expected returns used are: $H_1^1 = H_0 + 4db$, $H_1^2 = H_0 + 7dB$ and $H_1^3 = H_0 + 10dB$. The post-processed simulations of the avoidance commands do not include the path planner incorporated in the cost function, it is purely reactionary for avoiding collisions.

It was observed that when increasing the sensitivity settings, the algorithm exhibited a higher false alarm rate, primarily due to the increased detection of boat noise. This led to the detection rate improving as the algorithm became more sensitive to obstacles in the environment. Conversely, at lower sensitivity settings, the algorithm showed a lower false alarm rate, as it was less prone to detecting spurious features such as boat noise in the sonar data. However, this benefit came at the cost of a decreased detection rate, as the target was not detected immediately in the port and starboard beams. 

Figure \ref{Post Processed Path} illustrates the path followed by the AUV during the mission in Claytor Lake. In each subplot, the blue and red triangles represent port and starboard commands, respectively, pointed in the direction of the commanded maneuver. The starting position of the AUV is shown as a circle, the final position is an X and the target is shown as an asterisk.

To analyze the impact of sensitivity settings on the post-processed runs, Figure \ref{PPPA} presents a run with low sensitivity. This sensitivity setting was consistent with the setting used in the actual run. In the low sensitivity run, the AUV initiates a port command but corrects itself to a starboard command. It should be noted that although a port maneuver could have continued, the optimal trajectory from the start would have been a starboard maneuver. Some false alarms caused by boat noise can be observed in the figure after the AUV passes the target, which can be seen in the sonar returns in Figure \ref{ReturnsA} between 75 and 80 seconds.

In Figure \ref{PPPB}, another run is depicted with a medium sensitivity setting. Here, the AUV initiates a port command again but corrects to a starboard command more quickly in comparison to the low sensitivity case. However, the increased sensitivity lead to a higher number of false alarms caused by boat noise.

Figure \ref{PPPC} showcases the high sensitivity case. In this scenario, the AUV promptly begins with the correct starboard command, surpassing the medium and low sensitivity cases. However, there are additional false alarms at earlier stages after the AUV passes the target compared to the low and medium sensitivity cases.

Lastly, Figure \ref{PPPD} presents the actual run that took place. This run includes the additional costs incurred from the path planner as the AUV maneuvers away from the target, as discussed in Section \ref{CA3}. The maneuvers near the end of the run are a result of the additional costs, as the distance to the waypoint approaches and the AUV corrects itself towards the goal.

The results clearly demonstrate the trade-off between detection rate and false alarm rate with varying sensitivity settings. The algorithm's performance can be fine-tuned by adjusting the sensitivity based on the specific requirements of the application.

For an illustration of the commanded maneuvers given and the returns from the sonar, Figure \ref{Post Processed Returns} presents the returns of the center beam obtained during the run and the commanded maneuvers for each of the sensitivity levels. Figure \ref{ReturnsA} is over the entire leg of the lawnmower path, specifically illustrating the commands after maneuvering to avoid the obstacle, enabling the AUV to return to the designated waypoint. For a closer examination of the commands as the AUV approaches the obstacle, Figure \ref{ReturnsB} provides a zoomed-in view. In both figures, the results of the commanded maneuvers are shown below each of the sonar returns. The commanded maneuvers from the actual run is represented by the top row as the solid blue line. The subsequent rows depict the commands derived from the post-processed data where the commanded maneuvers from the low, medium and high sensitivity runs are depicted as a dashed red line, a dotted yellow line, and a dashed and dotted purple line, respectively. From the figure, it is clear the the high sensitivity case detects the obstacle the quickest and is less likely to switch back and forth between commanded maneuvers. Boat noise is detected in the starboard beam at approximately 65 seconds. After propagating the polar map, the boat noise results in false alarms, which do not occur in the medium or low sensitivity cases until around the 75 second mark.

An illustrative example of the AUV's polar map is shown in Figure \ref{localpost}, where a horizontal slice of the map is shown for each of the tested sensitivity levels. The beams are outlined in the figures, where the cells for each of the beams is between two of the red lines. The intensity of the return corresponds to the probability of an obstacle, where darker colors represent higher probability of obstacles. In each of the local polar maps, the target is 21 meters away from the AUV and the polar map is a snapshot of the 36 second mark corresponding to the 36 second mark in Figure \ref{field}. The low sensitivity case is shown in Figure \ref{local10}, where the target is not detected in any of the beams. In the medium sensitivity case from Figure \ref{local7}, the target is detected in the center beam. Figure \ref{local4} shows the high sensitivity case, where the target is detected in both the center and port beams. 

It is important to note that other factors, such as environmental conditions and obstacle characteristics, may also influence the algorithm's performance. Future research could explore the development of adaptive algorithms that dynamically adjust the sensitivity based on the real-time analysis of the sonar data and the environment.

The post-processed results indicate that increasing the sensitivity of the detection algorithm improves the detection rate at the expense of an increased false alarm rate. Conversely, decreasing the sensitivity reduces false alarms but may lead to delayed detection of obstacles. These findings provide valuable insights for optimizing the algorithm's performance in underwater obstacle detection applications.

\section{Conclusion}

A method for collision avoidance tightly coupled with obstacle detection and planning was presented and verified with field experimentation. The field experiments show the feasibility of a low-cost sonar for collision avoidance in small AUVs. Post-processed results demonstrate the effects of the chosen sensitivity levels, showing the importance of the sensitivity level on the detection and false alarm rates. Overall, we have demonstrated a novel sonar configuration for collision avoidance and verified it's usefulness with real-world data. 

\printcredits

\section*{Declaration of competing interest}

The authors declare that they have no known competing financial interests or personal relationships that could have appeared to influence the work reported in this paper.

% \section*{Data availability}

% Data not subject to the IP of the sonar developer will be made available on request.

\section*{Acknowledgments}
This work was supported by ATLAS North America. 

\bibliographystyle{cas-model2-names}

\bibliography{bibi}

\end{document}